\documentclass[runningheads]{llncs}
\usepackage[T1]{fontenc}
\usepackage{graphicx}
\usepackage{booktabs}
\usepackage[misc]{ifsym}
\newcommand{\corr}{(\Letter)}
\usepackage{mwe}
\usepackage{xcolor}
\usepackage{amsmath,amssymb,amsfonts}
\usepackage{subcaption}
\usepackage{tabularx}
\usepackage{algorithm}
\usepackage{algpseudocode}

\DeclareMathOperator*{\argmax}{arg\,max}

\newcommand{\net}{f}
\newcommand{\indim}{n}
\newcommand{\outdim}{m}
\newcommand{\weight}{\textbf{W}}
\newcommand{\bias}{\textbf{b}}
\newcommand{\preact}{\hat{z}}
\newcommand{\postact}{z}
\newcommand{\actfunc}{\sigma}
\newcommand{\inpoint}{x_0}
\newcommand{\outpoint}{y_0}

\newcommand{\concretelowersingle}{l}
\newcommand{\concreteuppersingle}{u}
\newcommand{\concretelowerindex}{l_{j}^{(i)}}
\newcommand{\concreteupperindex}{u_{j}^{(i)}}

\newcommand{\linlowerw}{\alpha_{L}}
\newcommand{\linlowerb}{\beta_{L}}
\newcommand{\linupperw}{\alpha_{U}}
\newcommand{\linupperb}{\beta_{U}}
\newcommand{\linupper}{h_{U}}
\newcommand{\linlower}{h_{L}}
\newcommand{\linupperindex}{h_{U,j}^{(i)}}
\newcommand{\linlowerindex}{h_{L,j}^{(i)}}
\newcommand{\indomain}{\mathcal{C}}
\newcommand{\segment}{\mathcal{S}}

\newcommand{\linlowernn}{\underline{g}}
\newcommand{\linuppernn}{\overline{g}}

\begin{document}

\title{Automated Design of Linear Bounding Functions\\for Sigmoidal Nonlinearities in Neural Networks}

\titlerunning{Automated Design of Linear Bounding Functions}
% If the full title of your paper is short enough to also fit in the running head, you can omit the abbreviated paper title here. You can check as follows: if you comment out the \titlerunning line, something will appear in the header of all odd-numbered pages of your PDF from page 3 onward. This something is either the full title (in which case all is well), or the error message "Title Suppressed Due to Excessive Length". If this error message appears, you're going to want to provide an abbreviated title within the \titlerunning command, because if you won't do it, Springer will do it for you.

%N.B.: Author information (both in the \author{} and \authorrunning{} command) should only be present in the Camera-Ready Version of your paper. The version that you initially submit for review, ought to be double-blind. So, when initially submitting your paper, use:
% \author{Author information scrubbed for double-blind reviewing}
\author{Matthias König\inst{1}\thanks{Work done while at University of Oxford.} \and
Xiyue Zhang\inst{2} \corr \and
Holger H. Hoos\inst{1,3} \and
Marta Kwiatkowska\inst{2} \and
Jan N. van Rijn\inst{1}
}
% You may leave out the orcidID information, if you want to.
% Use \corr to indicate the corresponding author. Note the spacing around the \corr command. Only one author can be the corresponding author.

%N.B.: comment out the \authorrunning{} command for the double-blind version of your paper submitted for review. Later, if your paper is accepted, use the command for the Camera-Ready Version.
\authorrunning{M. König et al.}
% If there are two authors, write 'A.L. Benjamin and C.C. Broadus Jr.'
% If there are more than two authors, '[...] et al.' is used.

\institute{Leiden University, The Netherlands \email{\{h.m.t.konig,j.n.van.rijn\}@liacs.leidenuniv.nl}
\and
University of Oxford, United Kingdom \email{\{xiyue.zhang,marta.kwiatkowska\}@cs.ox.ac.uk}
\and
RWTH Aachen University, Germany \\
\email{hh@aim.rwth-aachen.de}}

\maketitle              % typeset the header of the contribution

\begin{abstract}
The ubiquity of deep learning algorithms in various applications has amplified the need for assuring their 
robustness against small input perturbations such as those occurring in adversarial attacks. 
Existing \textit{complete} verification techniques offer provable guarantees for all robustness queries but struggle to scale beyond small neural networks.
To overcome this computational intractability, \textit{incomplete} verification methods often rely on convex relaxation to over-approximate the nonlinearities in neural networks.
Progress in tighter approximations has been achieved for piecewise linear functions.
However, robustness verification of neural networks for general activation functions (\emph{e.g.}, Sigmoid, Tanh) remains under-explored and poses new challenges.
Typically, these networks are verified using convex relaxation techniques, which involve computing linear upper and lower bounds of the nonlinear activation functions.
In this work, we propose a novel parameter search method to improve the quality of these linear approximations. 
Specifically, we show that using a simple search method, carefully adapted to the given verification problem through state-of-the-art algorithm configuration techniques, improves the average global lower bound by 25\% on average over the current state of the art on several commonly used local robustness verification benchmarks.

\keywords{Neural Network Verification \and Automated Algorithm Configuration \and Convex Relaxation}
\end{abstract}

\section{Introduction}

\begin{figure*}[t]
    \centering
    \begin{subfigure}{0.31\textwidth}
        \centering
        \includegraphics[width=\linewidth]{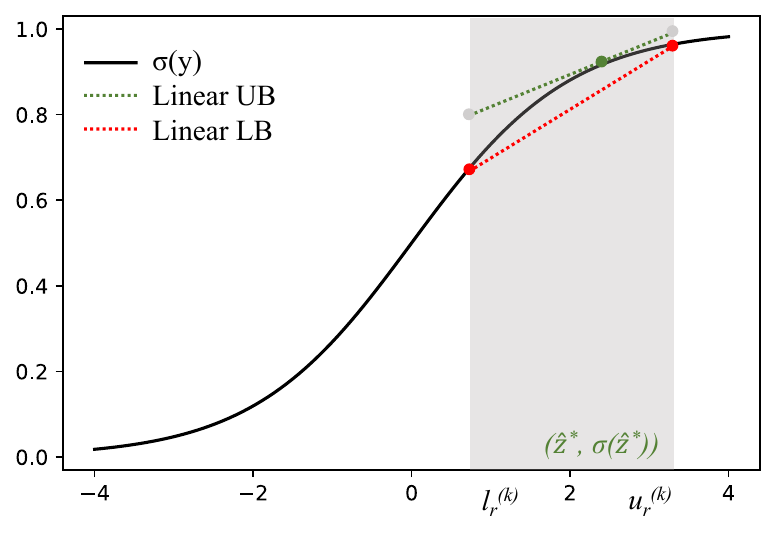}
        \caption{$\segment^{+}$}
        % \label{subfig5}
    \end{subfigure}
    \begin{subfigure}{0.31\textwidth}
        \centering
        \includegraphics[width=\linewidth]{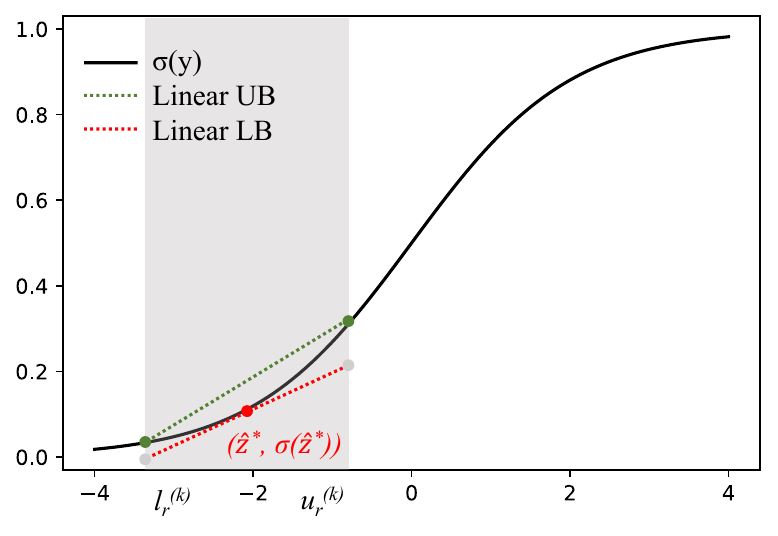}
        \caption{$\segment^{-}$}
        % \label{subfig7}
    \end{subfigure} 
    \begin{subfigure}{0.31\textwidth}
        \centering
        \includegraphics[width=\linewidth]{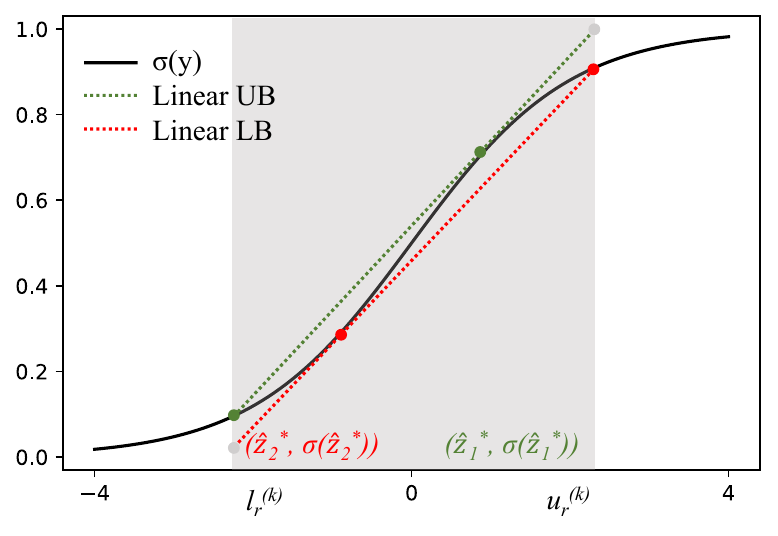}
        \caption{$\segment^{\pm}$}
        % \label{subfig12}
    \end{subfigure}
    \caption{Linear bounding rules for different cases of the Sigmoid activation function. The x-axis shows the pre-activation bounds, while the y-axis indicates the output of the activation function.}
    \label{fig-sigmoid}
\end{figure*}

Over the last decade, deep learning algorithms have gained increasing significance as essential tools across diverse application domains and usage scenarios. 
Their applications range from manoeuvre advisory systems in unmanned aircraft to face recognition in mobile phones (see, \emph{e.g.}, \cite{julian2019deep}).
Simultaneously, it is now widely acknowledged that neural networks are susceptible to adversarial attacks, where a given input is manipulated to cause misclassification \cite{szegedy2013intriguing}. 
Remarkably, in image recognition tasks, the perturbation required can be so subtle that it remains virtually imperceptible to the human eye.

Numerous methods have been proposed to 
evaluate the robustness of neural networks against adversarial attacks. 
Early methods involve empirical attacks \cite{goodfellow2014explaining,kurakin2016adversarial,carlini2017towards}; however, these approaches do not provide a comprehensive assessment of neural network robustness, as a defence mechanism against one type of attack might still be vulnerable to another, potentially novel, class of attacks. 
Consequently, there have been efforts to develop approaches that formally verify the robustness of neural networks against adversarial attacks \cite{katz2017reluplex,gehr2018ai2,xiang2018output,bunel2018unified,tjeng2017evaluating,botoeva2020efficient,HenriksenLomuscio20}. 
These formal verification methods enable a principled assessment of neural network robustness, providing provable guarantees on desirable properties, typically in the form of a pair of input-output specifications. 
However, \textit{complete} verification of neural networks, where the verifier is theoretically guaranteed to provide a definite answer to the property under verification, is a challenging NP-complete problem \cite{katz2017reluplex}. 
Solving this problem usually requires computationally expensive methods, \emph{e.g.}, SMT solvers~\cite{katz2017reluplex} or mixed integer linear programming systems~\cite{tjeng2019evaluating}, limiting scalability and efficiency. 

The aforementioned computational complexity is mainly due to the nonlinear activation functions in a neural network, 
which results in the neural network verification problem becoming a non-convex optimisation problem.
In light of this, \textit{incomplete} methods have been proposed that exploit \emph{convex relaxation} techniques, which approximate the nonlinearities using linear symbolic bounds, to provide sound and efficient verification~\cite{BakEtAl20,SinghEtAl19,zhangEtAl2018,gehr2018ai2,SinghEtAl18}; completeness can be achieved through branch and bound (see, \emph{e.g.} \cite{BunelEtAl18}).
Most formal verification methods are limited to ReLU-based networks (see \cite{li2020sok}), which satisfy the piecewise linear property. 
However, convex relaxation techniques are 
applicable to \emph{more general} commonly-used activation functions, such as 
Sigmoid or Tanh, by approximating them in terms of piecewise linear functions. 
This enables an extension of formal verification methods based on convex relaxation to general activation functions, though these remain under-explored.

An important application of neural network verification tools is \emph{robustness certification}, which computes guarantees that the prediction of the network is stable (invariant) around a given input point. 
$\mathrm{CROWN}$~\cite{zhangEtAl2018} is the first generic framework leveraging adaptive linear bounds to efficiently certify robustness for general activation functions.
Another series of techniques, \emph{e.g.}, $\mathrm{DeepZ}$~\cite{SinghEtAl18}, $\mathrm{DeepPoly}$~\cite{SinghEtAl19} employs abstract interpretation  
and abstract transformers for commonly-used activation functions, such as Sigmoid and Tanh.
\(\mathrm{CROWN}\) computes linear upper and lower bounds of nonlinear functions, while $\mathrm{DeepZ}$ and  $\mathrm{DeepPoly}$ incorporate convex relaxation into the abstract transformer.

Convex relaxation methods typically over-approximate the activation functions in all nonlinear neurons in the neural network, which inevitably introduces imprecision. 
Consequently, the corresponding verification algorithms may fail to prove the robustness of a neural network when the original network satisfies the specification and thus weaken certification guarantees. 
A globally, \emph{i.e.}, network-wise, tighter over-approximation is crucial for strong certification guarantees.
However, for Sigmoidal activations, it remains unclear how to design or configure the linear approximation of nonlinear neurons to achieve globally tight output bounds, which are directly used to determine the robustness of neural networks with respect to a specific property under investigation.

Motivated by the need to reduce imprecision 
of the bounding functions (as an over-approximation of the activation function), in this work we introduce an automated and systematic method to compute tighter bounding functions to improve certification guarantees. 
These bounding functions are defined by the tangent point, where they touch the activation function.
To this end, we propose a novel parameter search method for identifying the tangent points to find better-suited linear bounding functions by considering different cases of Sigmoidal activation functions.
To tackle the infinite search space (of tangent points) for the bounding functions, our approach leverages state-of-the-art algorithm configuration techniques.

Concretely, we use automated algorithm configuration techniques to find optimal hyper-parameters of the search method used for obtaining the tangent points of the linear bounding functions, which has previously been done using binary search \cite{zhangEtAl2018}.
These hyper-parameters control the initial tangent point per neuron as well as the rate at which these initial points are updated (we will refer to the latter as a \textit{multiplier} for the remainder of this work) until a feasible bounding function has been obtained.
Notice that these hyper-parameters are set for the entire network, \emph{i.e.}, all neurons share the same starting point and multiplier; however, they eventually result in different tangent points, as the search process only ends once a feasible bounding function has been found.

Moreover, we show that, by using our proposed method, we improve the average lower bound on the network output by 25\% on average across several verification benchmarks, and can certify robustness for instances that were previously unsolved. 

\section{Related Work}
In the following, we give some 
background on using convex relaxation for neural network robustness verification
and on automated algorithm configuration.

\subsection{Convex Relaxation for Neural Network Verification}
% [Notation]
We use 
% $\spec$ 
$\net: \mathbb{R}^{\indim} \to \mathbb{R}^{\outdim}$
to denote a neural network trained to make predictions on an $m$-class classification problem.
Let $\weight^{(i)}$ denote the weight matrix and $\bias^{(i)}$ denote
the bias for the $i$-th layer. 
% ($i \in \{1, \cdots, L\}$).
We use $\preact^{(i)}$ to denote
the pre-activation neuron values, and $\postact^{(i)}$ to denote
the post-activation ones, such that we have $\preact^{(i)} = \weight^{(i)} \postact^{(i-1)} + \bias^{(i)}$ (\emph{i.e.}, the post-activation values of layer $i-1$ are linearly weighted to become the pre-activation values of layer $i$). 
%\jvr{isn't this the other way around? i.e., $\postact^{(i)} = \weight^{(i)} \preact^{(i-1)} + \bias^{(i)}$???}
We use $\actfunc$ to denote the activation functions in intermediate layers, such that $\postact = \actfunc(\preact)$.
In this work, we focus on neural networks with sigmoidal activation functions,
including Sigmoid $\actfunc(x)=1/(1+e^{-x})$ and Tanh $\actfunc(x)=(e^{x}-e^{-x})/(e^{x}+e^{-x})$.
We use $f(x)$ to denote the neural network output for all classes and
 $f_j(x)$ to denote the output associated with class $j$ (with $1 \le j \le m$).
 % while $f(x)$ denotes 
The final decision is computed by taking the label with the greatest output value,  \emph{i.e.}, $\argmax_j f_j(x)$.
% JvR: Might not be the cleanest notation, but I felt like this was needed to make ensure proper usage further on. 
% Arctan $\actfunc(x)=tan^{-1}(x)$.

% [General idea]
\textbf{Neural network verification.}
The robustness to adversarial perturbations is one of the most important properties of neural networks, requiring that the predictions of a neural network should be preserved for a local input region $\indomain$, typically defined as a $l_p$ norm ball with a radius of $\epsilon$ around the original input $\inpoint$, in the following way:

\begin{equation}
\forall x \in \indomain, \,\argmax_j\net_j(x)=\argmax_j\net_j(\inpoint).
\end{equation}
For an input instance $\inpoint$ with ground-truth label $\outpoint$,
verifying the robustness property can then be transformed into proving that
% \begin{equation}
$\forall x \in \indomain, \net_{\outpoint}(x)-\net_{j}(x) \ge 0 $
% \end{equation}
for all $j \neq \outpoint$ where $j \in [1,m]$.

Given a verification problem, for example, to check whether the output constraint $\net_{\outpoint}(x)-\net_{j}(x)\ge 0$ is satisfied, we can append an additional layer at the end of the neural network, such that the output property can be merged into the new neural network function 
$g:\mathbb{R}^{\indim} \to \mathbb{R}$ where $ g(x) = \net_{\outpoint}(x)-\net_{j}(x)$.
% \jvr{I have a feeling we are cutting a corner here as $j$ can be multiple values in $f_j$, whereas $g$ seems to disregard this notion. I think we should replace $g$ in this formulation by $g_j$, and then later tell how these multiple $g_j$ properties lead to $g$ which we use further on.}
% \jvr{If we decide to not fix this, I would suggest adding along the lines of the following sentence (please double check):} 
% \xyc{Yes, the sentence correctly solved the notation consistency. For each verification query $g_j$, we do the same verification procudure with bound computation.}
While this should formally be done for each possible network output $j \neq y_0$, to simplify the notation we use a single $g$. %refrain from this and use a single $g$ for notation purposes.

In this way, the verification problem can be formulated canonically as follows, i.e., to prove or falsify:
\begin{equation}
    \forall x \in \indomain, g(x) \ge 0 
\end{equation}

One way to verify the property is to solve the optimisation problem $\min_{x\in \indomain}g(x)$.
However, due to the nonlinearity of the activation function $\actfunc$,
the neural network verification problem is NP-complete~\cite{katz2017reluplex}.
To address such intractability, state-of-the-art verification algorithms leverage convex relaxation to transform the verification problem into a convex optimisation problem.
\begin{definition}[Convex relaxation of activation functions]
For a nonlinear activation function $\actfunc(\preact)$ with pre-activation bounds $\preact \in [\concretelowersingle, \concreteuppersingle]$, convex relaxation relaxes the non-convex equality constraint $\postact = \actfunc(\preact)$ to two convex inequality constraints by computing the linear lower and upper bounding functions  $\linlower(\preact)=\linlowerw\preact+\linlowerb$ and $\linupper(\preact)=\linupperw\preact+\linupperb$, such that $\linlower(\preact)\le \actfunc(\preact) \le \linupper(\preact)$.
\end{definition}

In Figure~\ref{fig-sigmoid}, the pre-activation bounds are indicated by the grey-shaded area, $\linlower$ is illustrated by the red dotted line, whereas $\linupper$ is indicated by the green dotted line.
Notice that the parameters $\linlowerw,\linlowerb,\linupperw,\linupperb$ of the linear functions depend on the pre-activation bounds $\concretelowersingle$ and $\concreteuppersingle$.
With such convex inequality constraints, we can propagate the relaxations through layers using an efficient back-substitution procedure to compute the linear lower bounds (denoted by $\linlowernn$) and upper bounds (denoted by $\linuppernn$) for the neural network $g$.

\begin{definition}[Convex relaxation of neural networks]\label{def:convex_nn}
A convex relaxation of the neural network $g: \mathbb{R}^{\indim} \to \mathbb{R}$ 
% represented by the neural network 
over an input region $\indomain$, 
are two linear functions $\linlowernn$ and $\linuppernn$ such that 
$\linlowernn(x) \le g(x) \le \linuppernn(x)$
for all $x \in \indomain$.
% $\min_{x\in \indomain}\net_{L,j}(x) \le \min_{x\in \indomain}\net_{j}(x) \le \min_{x\in \indomain}\net_{U,j}(x)$ for a specified input region $\indomain$ on any label $j$.
\end{definition}

% One way to verify the property is to solve the optimisation problem $\min_{x\in \indomain}g(x)$.
% To address the NP-completeness brought by the non-convexity of neural networks,
Using convex relaxation,
% we can derive a valid linear lower bound for $g(x)$, denoted as $\linlowernn$, as shown in Definition~\ref{def:convex_nn}. 
% using convex relaxation 
% As a result, 
the verification  problem is then reduced to the following convex optimisation problem:
\begin{equation}
\label{eq:g}
g^{\ast}=\min_{x\in \indomain}\linlowernn(x)
\end{equation}
% $$\min_{x\in \indomain}(\net_{L,\outpoint}(x)-\net_{U,j}(x))$$
% For any label $j \in L\setminus{\outpoint}$, we can append a verification specification as an additional layer of the neural network, the problem can be formulated canonically into the following one-dimensional optimisation problem
% for any $j \in L \setminus \outpoint$:
% \begin{align*}
% g_j^{\ast} = \min_{x\in \indomain}g_{j}(x) =\min_{x\in \indomain}\net_{\outpoint}(x)-\net_{j}(x).
% \end{align*}
% \mk{Do we not need the L and U subscripts? Also, should we number the equations? Can we define g(x) somewhere in the previous paragraph?}
The robustness property is proved if the optimal solution  $g^{\ast} \ge 0$, as we have $\forall x \in \indomain, \, g(x) \ge \linlowernn(x)$.
Meanwhile, convex relaxation of the nonlinear constraints inevitably introduces approximation.
As a consequence, the computed minimum 
% \mkc{explain consequences - is it unknown/possibly unsafe?} 
might fail to satisfy $\min_{x \in \indomain} \linlowernn(x)\ge0$ even in cases in which the network is robust, \emph{i.e.}, $\min_{x \in \indomain} g(x)\ge0$. 
Instead, the true verification remains unknown.
% This could be a case of bound overestimation with fake counter-examples. 
Therefore, improving the quality of the approximation is crucial to reducing false failures in robustness verification, and thus strengthening certification guarantees, which is the aim of our contribution.

\label{sec: background}

\subsection{Automated Algorithm Configuration}

In general, the algorithm configuration problem can be described as follows: 
Given an algorithm \(A\) (also referred to as the \textit{target algorithm}) with parameter configuration space \(\Theta\), a set of problem instances \(\Pi\), and a cost metric \(c : \Theta \times \Pi \rightarrow \mathbb{R}\), find a configuration \(\theta^{\ast} \in \Theta\) that minimises cost \(c\) across the instances in \(\Pi\):

\begin{equation}
    \label{eq:smac}
    \theta^\ast \in \underset{\theta \in \Theta}{\mathrm{arg\,min}}\underset{\pi \in \Pi}{\sum}c(\theta,\pi)
\end{equation}

The general workflow of the algorithm configuration procedure starts with selecting a configuration \(\theta \in \Theta\) and an instance \(\pi \in \Pi\). 
Next, the configurator initialises a run of algorithm \(A\) with configuration \(\theta\) on instance \(\pi\), and measures the resulting cost \(c(\theta,\pi)\). 
The configurator uses this information about the target algorithm's performance to find a configuration that performs well on the training instances.
This is enabled by a surrogate model, which provides a posterior probability distribution that characterises the potential cost for $c(\theta, \pi)$ at a configuration $\theta$. 
At every time point $t$, we have a new observation of the cost $c_t$ at a new configuration point $\theta_t$.
The posterior distribution is updated based on the augmented observation set $S_t=S_{t-1} \cup \{(\theta_t, c_t)\}$, and the next configuration point is selected by maximising the acquisition function in the following form:
\begin{equation}
\theta_{t+1}=\underset{\theta \in \Theta}{\mathrm{arg\,max}} \, a_t(\theta, S_t)
\end{equation}
where $a_t$ denotes the acquisition function.
A typical choice for the acquisition functions is the expected improvement \cite{hutter2011sequential}.

Once the configuration budget (e.g., time budget or number of trials) is exhausted, the procedure returns the current incumbent \(\theta^\ast\), which represents the best configuration found so far. 
Finally, when running the target algorithm with configuration \(\theta^\ast\), it should result in lower cost (such as average running time) or improved solution quality across the benchmark set.

Automated algorithm configuration has already been shown to work effectively in the context of formal neural network verification \cite{KonEtAl22}, but also in other, related domains, such as SAT solving \cite{hutter2007boosting,hutter2017configurable}, scheduling \cite{chiarandini2008modular}, mixed-integer programming \cite{hutter2010automated,lopez2014automatically}, evolutionary algorithms \cite{bezerra2015automatic}, answer set solving \cite{gebser2011portfolio}, AI planning \cite{vallati2011automatic}, and machine learning \cite{thornton2013auto,feurer2015initializing}.

\section{Method}

We consider the task of local robustness verification with Sigmoidal activation functions.
To this end, we focus on convex relaxation based perturbation analysis as employed in the \(\mathrm{CROWN}\) framework \cite{zhangEtAl2018}, and use automated algorithm configuration techniques to improve the linear bounds of the nonlinear activation functions.
Notice that, by default, \(\mathrm{CROWN}\) employs binary search to obtain the points at which the linear bounding functions are tangent to the activation function \cite{zhangEtAl2018}.

In this study, on the other hand, we used \(\mathrm{SMAC}\) \cite{hutter2011sequential} to guide the search for suitable tangent points.
\(\mathrm{SMAC}\) is a widely known, freely available, state-of-the-art configurator based on sequential model-based optimisation (also known as Bayesian optimisation). 
The main idea of \(\mathrm{SMAC}\) is to construct and iteratively update a statistical model of target algorithm performance to guide the search for promising configurations; \(\mathrm{SMAC}\) uses a Random Forest regressor \cite{breiman2001random}.

\subsection{Configuration Objective}
The objective of the configuration procedure is to maximise the minimum of the global lower bound $g^\ast$ as defined in Equation~(\ref{eq:g}) for each instance under verification.
Notice that the optimisation procedure that solves Equation~(\ref{eq:g}) is performed separately for each instance, \emph{i.e.}, $|\Pi| = 1$.
Recall that an instance is verified to be robust when $g^{\ast}>0$,
as outlined in Section~\ref{sec: background}.

By design, \(\mathrm{SMAC}\) solves a minimisation problem; see Equation~(\ref{eq:smac}).
Since we are interested in maximising the lower bound of the network output, we apply appropriate sign changes and define the cost metric $c$ as the negative of the global lower bound.

\subsection{Configuration Space}

The \(\mathrm{CROWN}\) framework \cite{zhangEtAl2018} proposed a general certification solution for neural networks with nonlinear activation functions, which relies on convex relaxation to compute the output bounds.
Since Sigmoidal activation functions (Sigmoid/Tanh/Arctan) share the same features, that is, convex on the negative side ($x<0$) and concave on the other side ($x>0$), the authors of \cite{zhangEtAl2018} leveraged this feature and proposed a general method to compute the parameters (\emph{i.e.}, the tangent points) of linear upper and lower functions $\linupper$, $\linlower$. 

Based on the curvature of Sigmoidal activation functions, each nonlinear neuron of the $i$-th layer (denoted as $[n_{i}]$) is categorised into one of the three cases: $\segment^{+}$, $\segment^{-}$, and $\segment^{\pm}$ where 
$\segment^{+} = \{j \in [n_{i}] \mid 0\le \concretelowerindex \le \concreteupperindex\}$, 
$\segment^{-} = \{j \in [n_{i}] \mid \concretelowerindex \le \concreteupperindex \le 0\}$, 
and $\segment^{\pm} = \{j \in [n_{i}] \mid \concretelowerindex \le 0 \le \concreteupperindex \}$.
Intuitively, $\segment^{+}$ represents the case in which the pre-activation bounds are both positive, $\segment^{-}$ represents the case in which they are both negative and $\segment^{\pm}$ represents the case in which $\concretelowersingle$ is negative and $\concreteuppersingle$ is positive.
Different bounding rules are then proposed for these three cases; these are illustrated in Figure~\ref{fig-sigmoid}.

\subsubsection{Bounding rules for $S^+$ domain} 
For any node $j \in \segment^{+}$, the activation function $\actfunc(\preact_j)$ is concave; 
hence, a tangent line of $\actfunc(\preact_j)$ (represented in green in Figure~\ref{fig-sigmoid}) at any tangent point  $\preact^{\ast}_j \in [\concretelowerindex, \concreteupperindex]$ is a valid upper bounding function $\linupperindex$,
and the linear function passing the two endpoints, $(\concretelowerindex,\actfunc(\concretelowerindex))$ 
and $(\concreteupperindex,\actfunc(\concreteupperindex))$, serves as a valid lower bounding function $\linlowerindex$.

\begin{algorithm}[t!]
\caption{Our proposed tangent point search method for sigmoidal functions. It details only the search method in the $S^+$ domain (for space reasons). 
The hyper-parameter $s$ determines the initial tangent point and the hyper-parameter $\psi$ determines the rate by which $s$ is changed. 
The values of $s$ and the update rate $\psi$ are not exposed as hyper-parameters in vanilla \(\mathrm{CROWN}\).
In our case, these are optimised by \(\mathrm{SMAC}\), where \(\mathrm{SMAC}\) uses a fixed budget of $n_{\mathtt{max}}$ evaluation calls.
Notice that in vanilla \(\mathrm{CROWN}\), binary search is performed once initial bounds have been obtained to move tangent points closer to 0; this step is not performed in our method.
}
\label{algorithm}
\begin{algorithmic}[1]
\Procedure{OptimiseSearchParameters}{$x$, $s$, $\psi$}
    \State Initialize cost $c_0=\inf$ \Comment{first iteration of the optimisation loop can overwrite it}
    \State Initialise observation set $S_0 = \{\emptyset\}$
    \For {$n=1$, $\cdots$, $n_{\mathtt{max}}$}
    
    \State Select starting point and multiplier rate based on observation set
    \State $(s_n, \psi_n) \gets \underset{(s,\psi) \in \Theta}{\mathrm{arg\,max}} \, a_n((s,\psi), S_{n-1})$
    % \ForAll {Nodes in neural network $f$}
    \ForAll {node $j$ in network $f$}
    \If{$j \in S^+$}
    % \State Initialise upper bound tangent point for current node $\preact^{\ast} \gets s_n$
    \State Initialise upper bound tangent point for current node $\preact^{\ast}_j \gets s_n$
    \While {Tangent point $\preact^{\ast}_j$ leads to invalid upper bound $\linupperindex$}
        
        \State Update $\preact^{\ast}_j$: $\preact^{\ast}_j \gets \preact^{\ast}_j \cdot \psi_n$ 
    \EndWhile
    \ElsIf{$j \in S^-$}
      \State{Determine tangent point for lower bounding function, in a similar fashion as above}
    \Else \Comment{In this case, $j \in S^\pm$; }
      \State{Determine tangent point for both lower bounding and upper bounding function, in a similar fashion as above}
    \EndIf
    \EndFor
     \State Evaluate cost $c_n=-1 \cdot g^{\ast}(x)$
     \Comment{Linear bounding layer-by-layer via convex relaxation as per Def. \ref{def:convex_nn} and Eq. \ref{eq:g}}
     \State Augment observation set 
     $S_n=S_{n-1}\cup \bigl\{\bigl((s_n, \psi_n), c_n\bigr)\bigr\}$
     \State Train surrogate model using $S_n$
     \If {$c_n < c^{\ast}$}
       \State{$s^{\ast} \gets s_n$, $\psi^{\ast} \gets \psi_n$, $c^{\ast} \gets c_n$}
     \EndIf
    \EndFor
    
    \State \Return   $c^{\ast}$, ($s^{\ast},\psi^{\ast}$)
    % \State Initialise starting point for upper bound tangent point $s_{upper} \gets 1 \cdot s_{upper}$ 
    % \While {Tangent point $\preact^{\ast}$ leads to invalid upper bounding function $\linupperindex$}
    %     \State Update $\preact^{\ast}$: $\preact^{\ast} \gets \preact^{\ast} \cdot \psi_{upper}$ 
    % \EndWhile
\EndProcedure
\end{algorithmic}
\end{algorithm}

To select good values of $\preact^{\ast}_j$, we propose a search method whose behaviour depends on two hyper-parameters, which are optimised by SMAC: 
\begin{itemize}
    \item starting point $s$ at which each tangent point $\preact^{\ast}_j$ is initialised; 
    \item multiplier $\psi$ to change the value of $\preact^{\ast}_j$ if the linear bound is found to be invalid. 
\end{itemize}
This search method is illustrated in Algorithm~\ref{algorithm}.
The algorithm initiates and maintains the best configuration $\theta^{\ast}=(s^{\ast}, \psi^{\ast})$ found so far, which leads to a tighter bounding function (defined by the tangent points $\preact^{\ast}_j$ for each node $j$) and, thus, a tighter global lower bound $g^{\ast}$. 
To find optimal settings for the hyper-parameters $s$ and $\psi$, SMAC samples values of $s$ and $\psi$ from a pre-defined configuration space $\Theta$. 
Configurations $\theta$, as introduced in Equation~(\ref{eq:smac}), are sampled from $s \in [0.01, 2]$ and $\psi \in [1.01, 3]$.
Note that these ranges are based on empirical observations; values outside this range typically result in extremely loose bounds.
In principle, other ranges could be used to sample values of $s$ and $\psi$, respectively, as long as $s>0$ and $\psi>1$.
Also note that the search method will be the same for the entire network; however, it will result in different tangent points per node. 

Based on a given hyperparameter configuration, we can identify the tangent points $\preact^{\ast}_j$ for each node $j$, and compute the global lower bound $g^\ast$ as well as the cost value $c=-1 \cdot g^\ast(x)$. 
At each iteration, we can acquire the parameter configuration based on the updated surrogate model (trained with the new observation set $S_n$).
With the selected candidate configuration, we perform a sanity check to ensure the validity of the linear upper bounding function incurred by the tangent points $\preact^{\ast}_j$ at each node.  
An upper bound is considered invalid if, at any given point $\preact^{\ast}_j$, the value of $\linupperindex(\preact^{\ast}_j)$ is smaller than the value of the nonlinear sigmoidal $\actfunc(\preact^{\ast}_j)$.
The global lower bound $g_n$ (also the cost value $c_n$) is then updated with the new upper bounding function.
The loop terminates and returns the best-achieved cost value $c^{\ast}$ as well as the corresponding configuration $(s^{\ast},\psi^{\ast})$ when the maximum iteration limit $n_{\mathtt{max}}$ is reached.
Notice that SMAC has several additional options that can slightly deviate from the description~\cite{hutter2011sequential}. 

While Algorithm~\ref{algorithm} elaborates on tangent point search for the upper bounding function, it can be similarly configured to search for lower bounding functions with two key modifications. Firstly, the parameters are initialised to different value ranges. Secondly, the validity evaluation is to check the bounding function will always return a lower value than the sigmoidal function. We introduce the details for $S^-$ and $S^\pm$ domains in the following.

\subsubsection{Bounding rules for $S^-$ domain} Symmetrically, for any node $j \in \segment^{-}$, $\linupperindex$ is defined as the linear function passing the two endpoints and 
$\linlowerindex$ can be a tangent line of $\actfunc(\preact_j)$ at any point $\preact^{\ast}_j \in [\concretelowerindex, \concreteupperindex]$.
In this case, we employ a similar search method, except that a configuration $\theta^*$ is now sampled from $s \in [-0.01, -2]$ and $\psi \in [-1.01, -3]$.
Again, other ranges could be used to sample values of $s$ and $\psi$, respectively, as long as $s<0$ and $\psi<-1$.
Furthermore, a bound is considered invalid if, at any given point $\preact^{\ast}_j$, the value of $\linlowerindex(\preact^{\ast}_j)$ is larger than the value of $\actfunc(\preact^{\ast}_j)$.

\subsubsection{Bounding rules for $S^\pm$ domain} 
Lastly, for any node $j \in \segment^{\pm}$, $\linupperindex$ is a tangent line passing $(\concretelowerindex,\actfunc(\concretelowerindex))$ and a tangent point $(\preact^{\ast}_1, \actfunc(\preact^{\ast}_1))$, where $\preact^{\ast}_1 \ge 0$, and $\linlowerindex$ is a tangent line that passes $(\concreteupperindex,\actfunc(\concreteupperindex))$ and a tangent point $(\preact^{\ast}_2, \actfunc(\preact^{\ast}_2))$,  where $\preact^{\ast}_2 \le 0$.
Again, we employ the same search method to obtain $\preact^{\ast}_j$ for each bound.
Moreover, for any $\linupperindex$, a configuration $\theta^*$ is sampled from $s \in [0.01, 2]$ and $\psi \in [1.01, 3]$, while for any $\linlowerindex$, a configuration $\theta^*$ is sampled from $s \in [-0.01, -2]$ and $\psi \in [-1.01, -3]$.
Notice that we can use similar rules for bounding Tanh functions, which share the Sigmoidal shape but range from -1 to 1 instead of 0 to 1.

% \afterpage{
\begin{table*}[t]
\caption{Experimental results obtained from our automated bound configuration method, compared to the vanilla \(\mathrm{CROWN}\) algorithm, which relies on binary search to obtain the tangent points (with $\alpha$-optimisation for Sigmoid networks). 
Boldfaced values indicate superior performance.
}
\begin{center}
\resizebox{\textwidth}{!}{
\begin{tabular}{|l|l|l|r|r|r|r|r|r|}
\hline
 & & & & \multicolumn{3}{c|}{\textbf{Avg. Global Lower Bound} $\mathbf{g^{\ast}}$} & \multicolumn{2}{c|}{\textbf{\# Certified Instances}} \\
\cline{5-9} 
 % & & & CROWN & CROWN & & CROWN & CROWN \\
\textbf{Dataset} & \textbf{Network} & \textbf{Activation} & \textbf{Epsilon} & Baseline & Configured & Improvement & Baseline & Configured \\
\hline
CIFAR10 & ConvMed & Sigmoid & 0.0313 & -15.611 & \bf-5.506 & 184\% & 0 & \bf1 \\
CIFAR10 & ConvMed & Sigmoid & 0.0157 & -1.727 & \bf-1.633 & 6\% & 13 & 13 \\
CIFAR10 & ConvMed & Sigmoid & 0.0078 & -0.709 & \bf-0.702 & 1\% & 29 & \bf31 \\
CIFAR10 & ConvMed & Sigmoid & 0.0039 & -0.440 & -0.440 & - & 39 & 39 \\
\hline
CIFAR10 & FNN\_6x500 & Sigmoid & 0.0313 & -18.448 & \bf-16.433 & 12\% & 0 & 0 \\
CIFAR10 & FNN\_6x500 & Sigmoid & 0.0157 & -17.819 & \bf-16.234 & 10\% & 0 & 0 \\
CIFAR10 & FNN\_6x500 & Sigmoid & 0.0078 & -15.075 & \bf-13.478 & 12\% & 0 & 0 \\
CIFAR10 & FNN\_6x500 & Sigmoid & 0.0039 & -6.731 & \bf-5.915 & 14\% & 5 & \bf8 \\
\hline
MNIST & ConvMed & Sigmoid & 0.3 & -5.153 & \bf-4.175 & 23\% & 3 & \bf4 \\
MNIST & ConvMed & Sigmoid & 0.12 & -9.484 & \bf-8.563 & 10\% & 0 & 0 \\
MNIST & ConvMed & Sigmoid & 0.06 & -0.110 & \bf-0.070 & 57\% & 54 & 54 \\
MNIST & ConvMed & Sigmoid & 0.03 & 2.753 & 2.756 & - & 92 & 92 \\
\hline
CIFAR10 & ConvMed & Tanh & 0.0313 & -65.849 & \bf-55.565 & 19\% & 0 & 0 \\
CIFAR10 & ConvMed & Tanh & 0.0157 & -44.504 & \bf-33.723 & 32\% & 0 & 0 \\
CIFAR10 & ConvMed & Tanh & 0.0078 & -10.835 & \bf-9.348 & 16\% & 2 & 2 \\
CIFAR10 & ConvMed & Tanh & 0.0039 & -2.299 & \bf-2.259 & 2\% & 15 & \bf16 \\
\hline
% \multicolumn{4}{l}{$^{\mathrm{a}}$Sample of a Table footnote.}
\end{tabular}}
\label{tab1}
\end{center}
\end{table*}
% }

\section{Setup for Empirical Evaluation}

We will empirically investigate the effectiveness of the proposed search procedure.
We consider two types of neural network architecture: convolutional neural networks (CNNs) and fully connected neural networks (FNNs).
We evaluate the effectiveness of our approach on the $\mathrm{ERAN}$ benchmark~\cite{muller2021scaling,SinghEtAl19a,SinghEtAl19,SinghGehr19,SinghEtAl18}.
Following the naming conventions, we refer to these networks as ConvMed and FNN\_6x500.
These networks are adversarially trained on the CIFAR-10 dataset with $\epsilon=0.0313$.
In addition, to investigate whether our findings hold for additional datasets, we consider a CNN with similar architecture that is adversarially trained on the MNIST dataset with $\epsilon=0.3$.
Furthermore, to demonstrate the generality of our method to other Sigmoidal activation functions, we perform the evaluation on a neural network trained on CIFAR-10 with similar architecture, adversarially trained with $\epsilon=0.0313$, but using Tanh activation functions.
Further details about the considered networks can be found in the ERAN repository.

We verify each network for local robustness with respect to the first 100 instances in the test set of the MNIST and CIFAR-10 datasets, respectively, initially with perturbation radii equivalent to the values used during training.
We also evaluate our method on a broader range of perturbation radii; specifically, we considered $\epsilon \in \{0.0157, 0.0078, 0.0039\}$ for CIFAR-10 and $\epsilon \in \{0.12, 0.06, 0.03\}$ for MNIST.
Notice that these values of $\epsilon$ are in line with commonly chosen values from the verification literature \cite{konig2024critically,wang2021beta,singh2019abstract}.

For verification, we employ \(\mathrm{CROWN}\) with default settings as provided by the authors. 
For Sigmoid-based networks, \(\mathrm{CROWN}\) also performs $\alpha$-optimisation \cite{DBLP:conf/iclr/XuZ0WJLH21}; however, this is not implemented for Tanh activation functions, where we run \(\mathrm{CROWN}\) without further optimisation.
As we are purely interested in the global lower bound on the network output, we skip the PGD attack used for upper bound computation.
For the configuration procedure, we set the number of evaluation calls performed by SMAC to 150, \emph{i.e.}, the configuration process terminates after 150 trials. 

\begin{figure*}[t!]
    \centering
    \begin{subfigure}{0.45\textwidth}
        \centering
        \includegraphics[width=\linewidth]{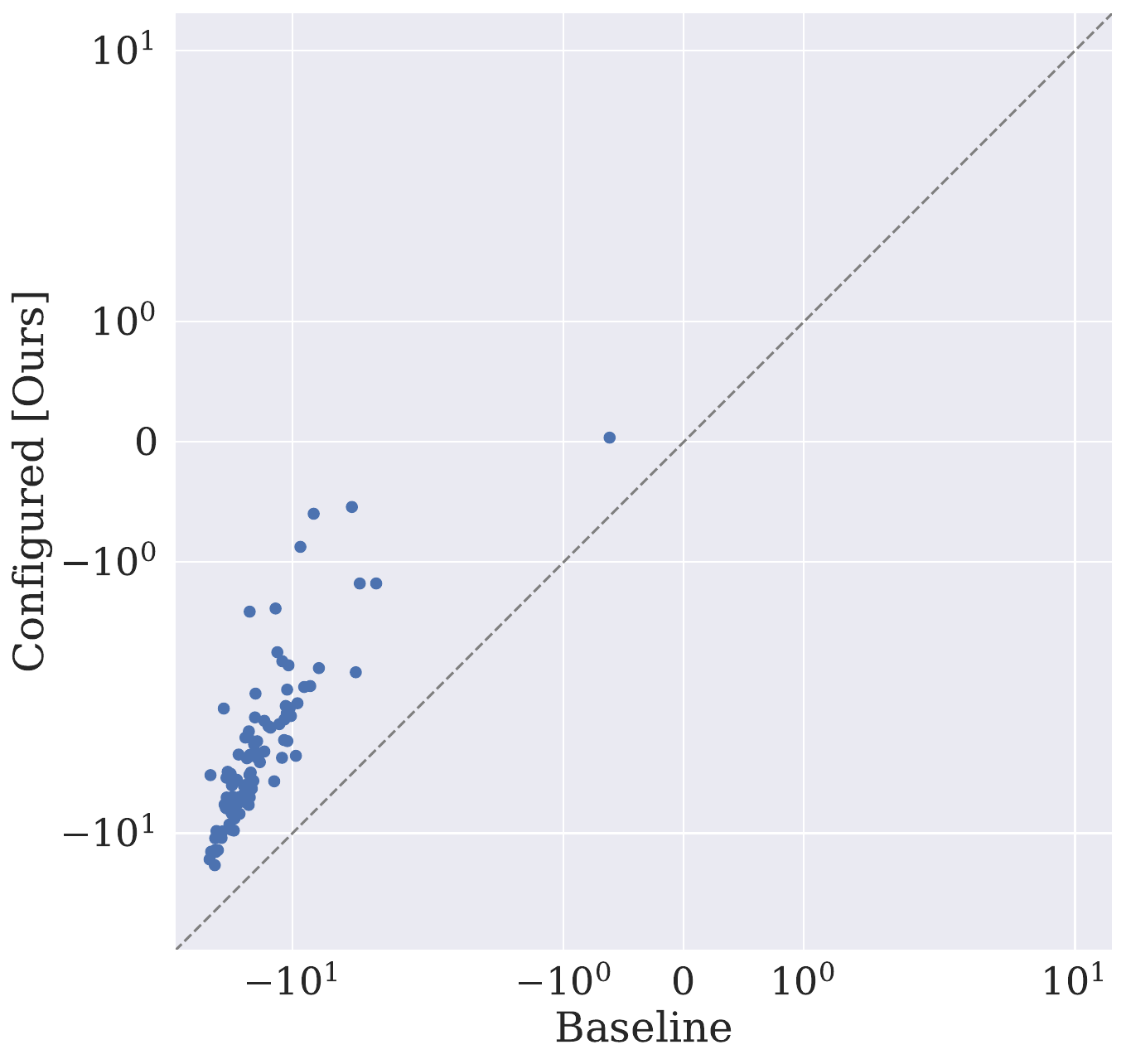}
        \caption{CIFAR10, CNN, \\$\epsilon=0.0313$}
        \label{subfig5}
    \end{subfigure}
        \centering
    \begin{subfigure}{0.45\textwidth}
        \centering
        \includegraphics[width=\linewidth]{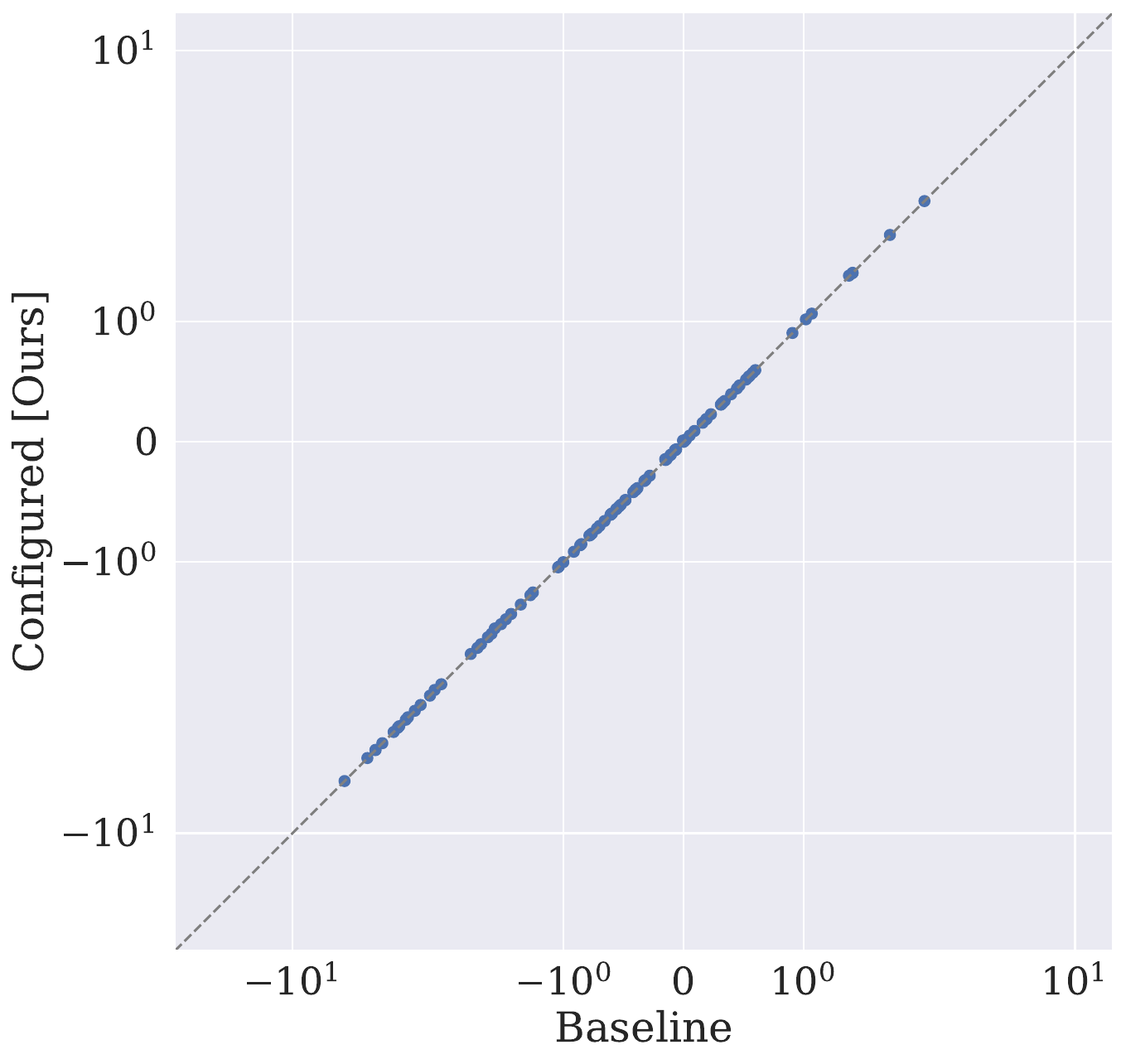}
        \caption{CIFAR10, CNN, \\$\epsilon=0.0078$}
        \label{subfig7}
    \end{subfigure} 
    \begin{subfigure}{0.45\textwidth}
        \centering
        \includegraphics[width=\linewidth]{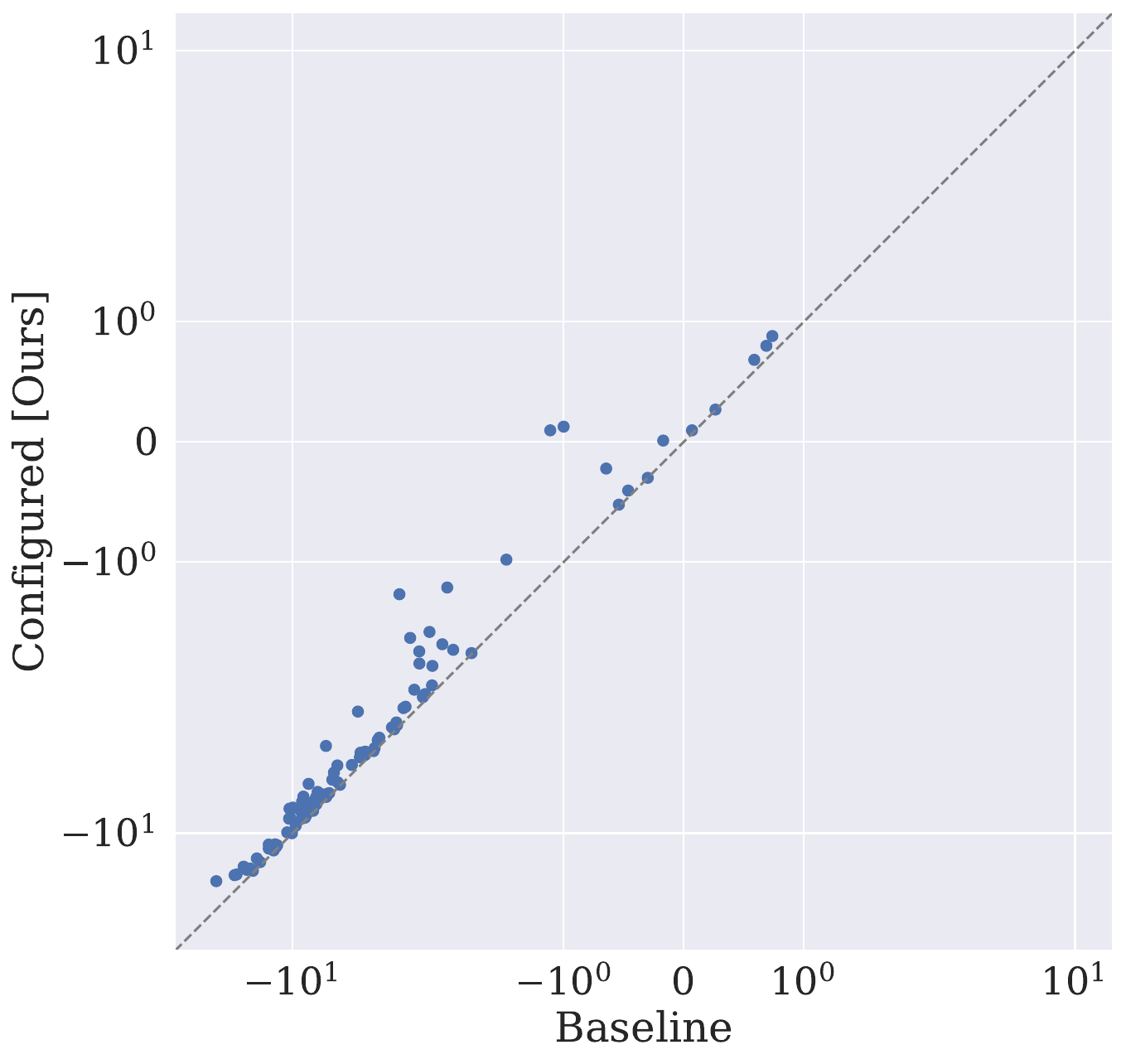}
        \caption{CIFAR10, FNN, \\$\epsilon=0.0039$}
        \label{subfig12}
    \end{subfigure}
    \begin{subfigure}{0.45\textwidth}
        \centering
        \includegraphics[width=\linewidth]{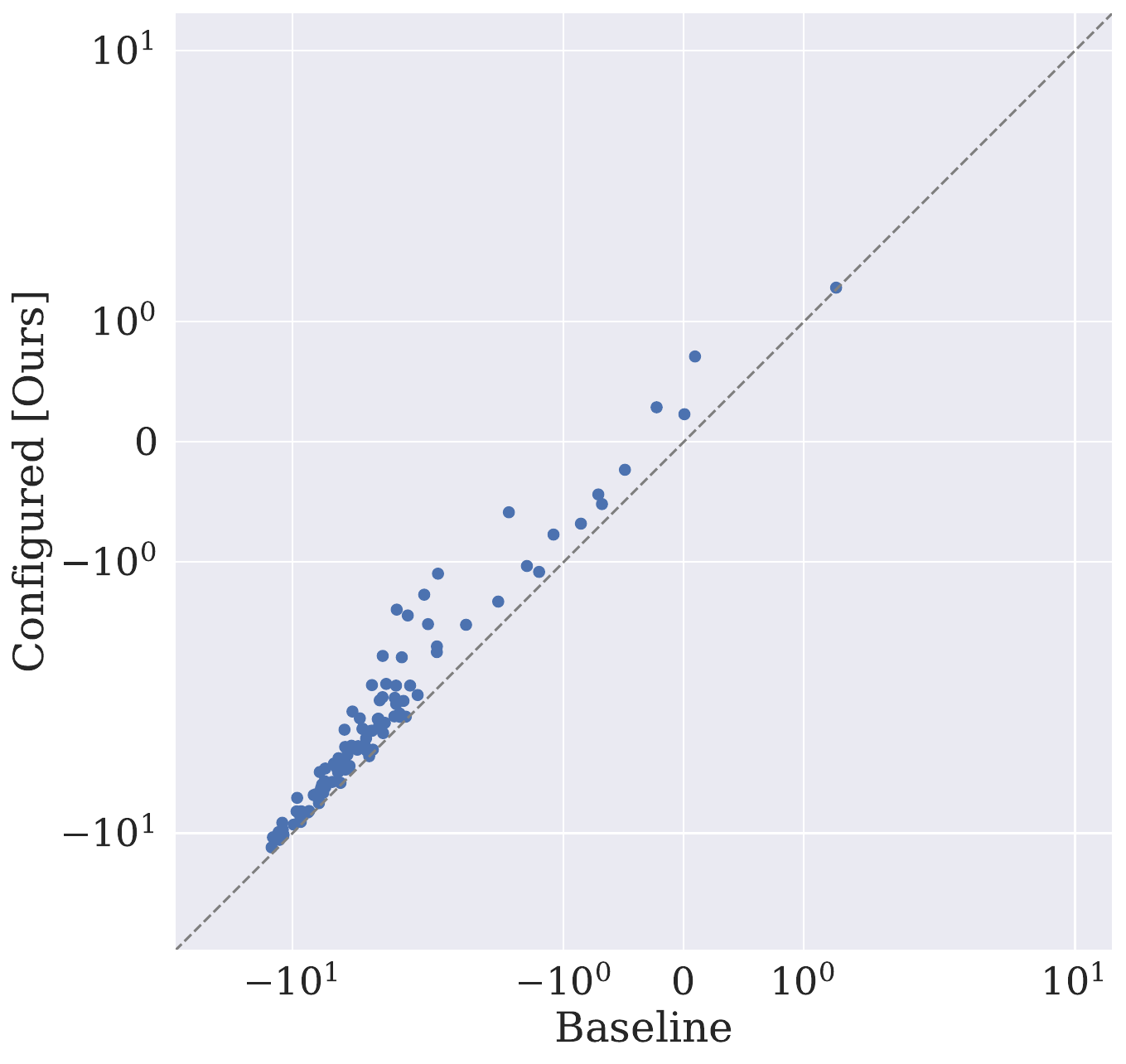}
        \caption{MNIST, CNN, \\$\epsilon=0.3$}
        \label{subfig1}
    \end{subfigure}
    \caption{Experimental results obtained for Sigmoid-based networks. 
    Each dot represents a problem instance and the global lower bound, \emph{i.e.}, the value of $g^{\ast}$, for that instance achieved by the baseline approach (x-axis) \emph{vs} our method (y-axis). 
    }
    \label{fig}
\end{figure*}

All experiments are performed on a cluster of machines equipped with NVIDIA GeForce GTX 1080 Ti GPUs with 11 GB video memory.

\section{Experimental Results and Discussion}

\begin{figure*}
    \centering
    \begin{subfigure}{0.31\textwidth}
        \centering
        \includegraphics[width=\linewidth]{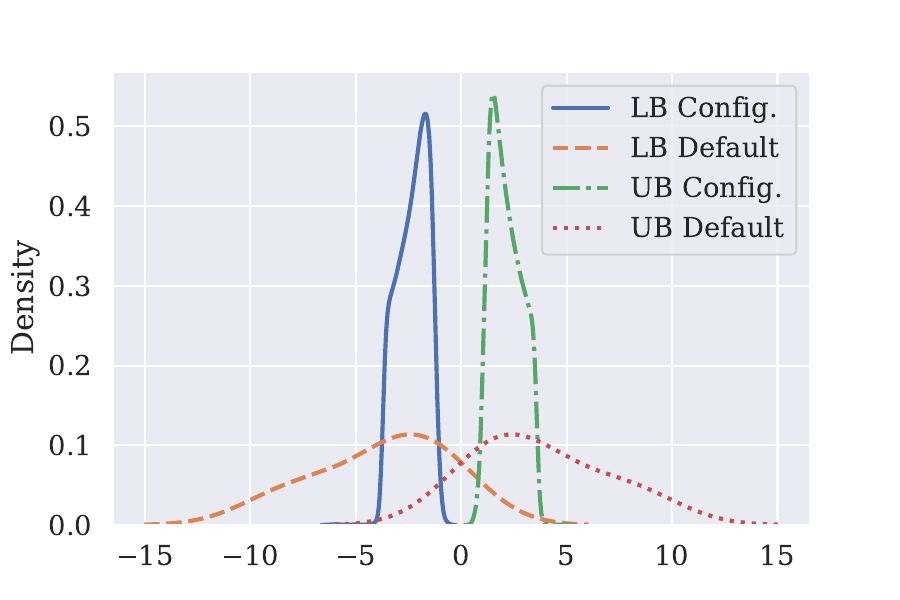}
        \caption{Layer 1}
        % \label{subfig5}
    \end{subfigure}
    \begin{subfigure}{0.31\textwidth}
        \centering
        \includegraphics[width=\linewidth]{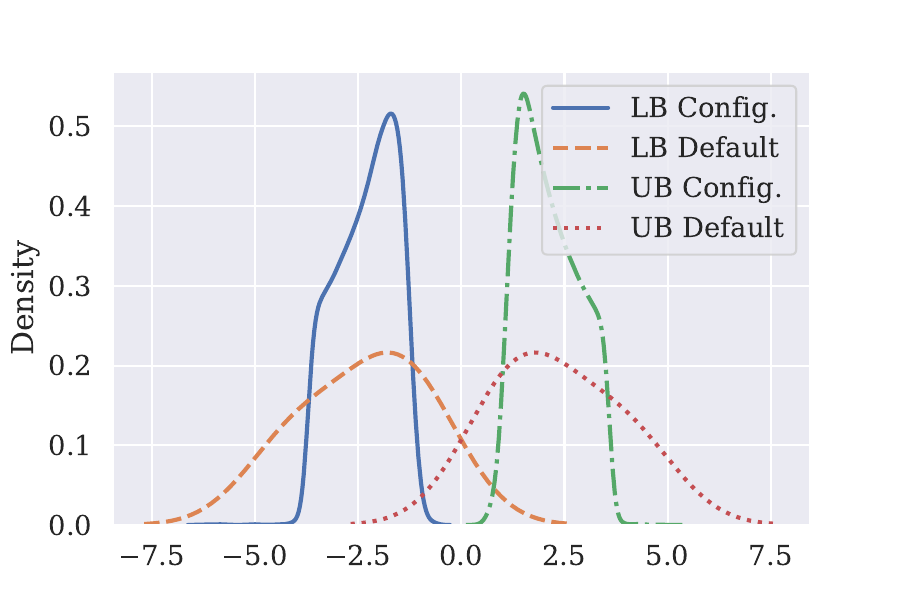}
        \caption{Layer 2}
        % \label{subfig7}
    \end{subfigure} 
    \begin{subfigure}{0.31\textwidth}
        \centering
        \includegraphics[width=\linewidth]{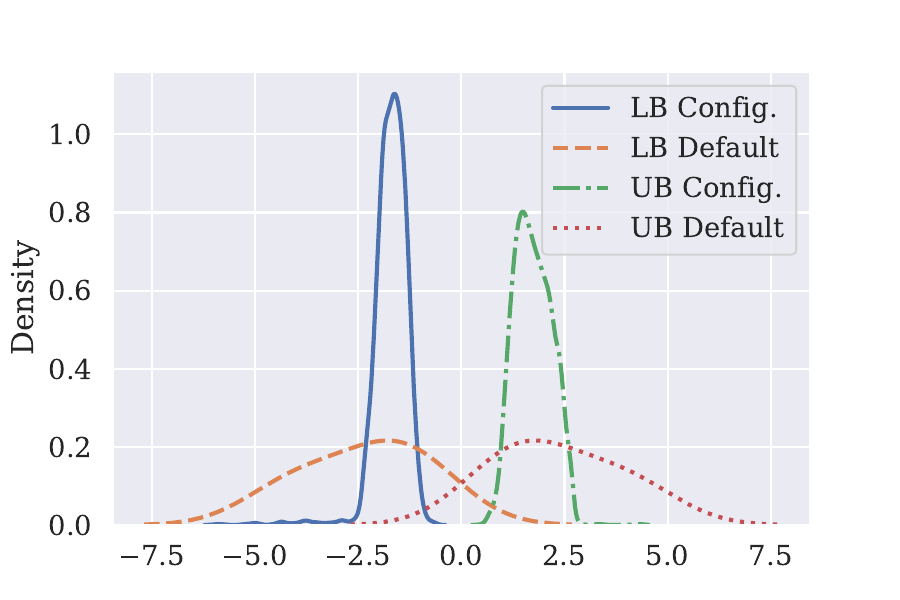}
        \caption{Layer 3}
        % \label{subfig12}
    \end{subfigure} \\
    \caption{Probability density functions of the values of $\preact^{\ast}_j$ obtained by our method as well as vanilla \(\mathrm{CROWN}\) for lower and upper bounding functions per activation layer. 
    Remember that $\preact^{\ast}_j$ determines the tangent point of the bounding function of a given node $j$.}
    \label{fig-dist}
\end{figure*}

Table~\ref{tab1} shows the average global lower bound over the given test instances achieved by the vanilla \(\mathrm{CROWN}\) algorithm, which represents our baseline, and those achieved by \(\mathrm{CROWN}\) in combination with our configured bounding method. 
In addition, we report the absolute number of instances for which robustness certification could be obtained by each approach.

\subsection{Sigmoid-based Networks}
We first report the results obtained for the Sigmoid-based CNN trained on the CIFAR-10 dataset.
These can be found in Table~\ref{tab1}.
When certifying this network with $\epsilon=0.03$, we obtained an improvement in the average global lower bound of 184\% (-15.611 \emph{vs} -5.506). 
This is also visualised in Figure~\ref{subfig5}. 
As shown there, the global lower bound is improved by almost an order of magnitude for some of the instances.
Furthermore, using our configured bounds, we could certify robustness for an instance that could not be solved by the baseline approach.

For $\epsilon=0.0157$ and $\epsilon=0.0078$, we achieved improvements of 6\% and 1\%, respectively.
Interestingly, although the latter is a rather small improvement in the average global lower bound, our method could verify two additional instances, which the baseline method was unable to solve.
These results are visualised in Figure~\ref{subfig7}.
Although instances generally lie very close to the equality line, our method could improve on instances with lower bounds very close to 0, for which even a very small increase can lead to certified robustness.

Lastly, when $\epsilon=0.0039$, we achieved a similar performance as the baseline method. 
This indicates that the effectiveness of configured bounds decreases as the perturbation radius becomes very small.

Next, we investigate whether our approach extends to fully connected neural networks; experimental results are shown in Table~\ref{tab1}.
In general, bounds obtained for this network type are looser than those obtained for the CNN, irrespective of the perturbation radius.
Nonetheless, our method achieved improvements in the average global lower bound between 10 and 14 per cent across all considered perturbation radii.

For FNNs, we achieved the greatest improvement when the perturbation radius is smallest, \emph{i.e.} $\epsilon=0.0039$. 
In this scenario, we were able to certify robustness for 3 additional instances, which were previously unsolved; see also Figure~\ref{subfig12} for a visualisation of these results.

Table~\ref{tab1} also shows the results from our experiments on the CNN trained on the MNIST dataset.
When verifying this network with $\epsilon=0.3$, we achieved an improvement in the average global lower bound of 23\%.
Furthermore, using our configured bounds, we could again certify robustness for an instance that could not be solved previously.
This is also visualised in Figure~\ref{subfig1}. 

For $\epsilon=0.12$ and $\epsilon=0.06$, we achieved improvements in the average global lower bound of 10\% and 57\%, respectively.
Lastly, when $\epsilon=0.03$, our method did not improve over the baseline.
This again shows that, for CNNs, configuring the bounds is less effective if the perturbation radius is minimal.

\subsection{Tanh-based Networks}

Next, we report the results obtained for the Tanh-based CNN trained on the CIFAR-10 dataset; these are also presented in Table~\ref{tab1}.
Notably, we achieved consistent improvement for any given value of $\epsilon$. 
Furthermore, we found that the global lower bounds are generally much lower than those obtained for the Sigmoid-based CNN, although verifying the same properties. 
Nevertheless, when $\epsilon=0.0039$, our method enables the verification of an additional instance, which was previously unsolved.
Overall, our results demonstrate the strength of our approach, and its potential to improve verification performance on networks with non-piecewise linear activation functions in general.

\subsection{Distribution of Tangent Points}

To gain a better understanding of the difference between bounding functions obtained by our method and vanilla \(\mathrm{CROWN}\), we perform an empirical analysis of tangent point distributions obtained for the linear bounding functions of the Sigmoid-based CNN network trained on CIFAR-10 when verified for local robustness with $\epsilon=0.0313$. 
Notice that this benchmark shows the greatest improvement in the average global lower bound. 

Figure~\ref{fig-dist} shows probability density functions of tangent points for lower and upper bounding functions per activation layer of the considered network. 
Notably, they show that the distribution of tangent points found by the configured bounding algorithm is much more centred around a specific value than those obtained by the baseline approach. 
Furthermore, the difference between the empirical distribution functions of tangent points obtained by our search method and the baseline approach was determined as statistically significant by means of a Kolmogorov–Smirnov test with a standard significance threshold of 0.05. 
Moreover, these empirical observations hint towards the existence of an optimal region for bounding parameters of Sigmoidal activation functions, which might be difficult to identify using the baseline approach without automated configuration of the tangent point search method.

\section{Conclusions and Future Work}

In this work, we have shown that automated algorithm configuration can provide a systematic and effective way for designing bounding functions of nonlinearities beyond the commonly studied piecewise linear activation functions (\emph{e.g.}, ReLU).
Specifically, our new method achieved consistent improvements in average global lower bound across several robustness verification benchmarks and perturbation radii. 

At the same time, we see several fruitful avenues for future work. 
First of all, the proposed search method for obtaining the tangent points is controlled by only two hyper-parameters.
A more sophisticated method, allowing for a more fine-grained configuration of the search method, could improve performance further.
In addition, one could configure the hyper-parameters of the search method as well as those of the $\alpha$-optimisation method, \emph{e.g.}, the learning rate of the projected gradient algorithm, jointly.
Overall, we see this study as a promising step towards the automated design of versatile and efficient neural network verification algorithms.

\subsubsection{Acknowledgment}

This research was partially supported by TAILOR, a project funded by EU Horizon 2020 research and innovation program under GA No. 952215. 
It further received funding from the ERC under the European Union’s Horizon 2020 research and innovation program (FUN2MODEL,
grant agreement No. 834115) and ELSA: European Lighthouse on Secure and
Safe AI project (grant agreement No. 101070617 under UK guarantee).

\bibliographystyle{splncs04}
\bibliography{references} 

\end{document}